\documentclass[12pt]{cmuthesis}

\usepackage{times}
\usepackage{fullpage}
\usepackage{graphicx}
\usepackage{amsmath}
\usepackage[numbers,sort]{natbib}
\usepackage[hyphens]{url}
\usepackage{hyperref}
\hypersetup{pageanchor=true,plainpages=false,bookmarksnumbered,breaklinks=true,
}
\usepackage[hyphenbreaks]{breakurl}
\usepackage{subcaption}
\usepackage{algpseudocode}
\usepackage{algorithm}
\usepackage{appendix}
\usepackage[final]{changes}
\usepackage{mdframed}
\usepackage{tikz}
\usetikzlibrary{arrows.meta}
\usetikzlibrary{decorations.pathmorphing}

\tikzset{
  is-a-arrow/.style={arrows={-{Latex[length=3mm]}}, very thick},
  x-y-z-arrow/.style={arrows={-{Latex[length=3mm]}}, very thick, dashed},
  res-is-a-arrow/.style={arrows={-{Latex[length=3mm]}}, double, double distance=2},
  res-x-y-z-arrow/.style={arrows={-{Latex[length=3mm]}}, double, double distance=2, dashed},
}

\usepackage[letterpaper,twoside,vscale=.8,hscale=.75,nomarginpar,hmarginratio=1:1]{geometry}

\begin {document} 
\frontmatter

\pagestyle{empty}

\title{ 
{\bf Score: A Rule Engine for the Scone Knowledge Base System}}
\author{\bf Jeffrey Chen}
\date{May 2021}
\Year{2021}
\trnumber{CMU-CS-21-116}

\committee{
Scott E. Fahlman, Carnegie Mellon University\\
Computer Science Department and Language Technologies Institute \\\quad\\
Alessandro Oltramari\\
Bosch Research \& Technology Center, Pittsburgh
}

\support{}
\disclaimer{}



\maketitle

\pagestyle{plain} 


\begin{abstract}
We present Score, a rule engine designed and implemented for the Scone knowledge base system. Scone is a knowledge base system designed for storing and manipulating rich representations of general knowledge in symbolic form. It represents knowledge in the form of nodes and links in a network structure, and it can perform basic inference about the relationships between different elements efficiently. On its own, Scone acts as a sort of ``smart memory" that can interface with other software systems. One area of improvement for Scone is how useful it can be in supplying knowledge to an intelligent agent that can use the knowledge to perform actions and update the knowledge base with its observations.

We augment the Scone system with a production rule engine that\deleted{ would} automatically perform\added{s} simple inference based on existing and \added{newly-}added \replaced{structures in Scone's knowledge base}{knowledge},\added{ potentially} improving the capabilities of any planning systems built on top of \replaced{Scone}{it}. Production rule systems consist of ``if-then" production rules that try to match their predicates to existing knowledge and fire their actions when their predicates are satisfied. We propose two kinds of production rules, if-added and if-needed rules, that differ in how they are checked and fired to cover multiple use cases. We then implement methods to efficiently check and fire these rules in a large knowledge base. The new rule engine is not meant to be a complex stand-alone planner, so we discuss how it fits into the context of Scone and future work on planning systems.
\end{abstract}

\begin{acknowledgments}
I would like to thank Professor Scott Fahlman for introducing me to Scone and providing me with guidance on how a rule engine in Scone should be designed. His insights on the overall structure of Scone as well as on other related work in production systems were instrumental in helping me write this thesis. I would also like to thank Alessandro Oltramari for agreeing to take the time to be on the thesis committee and for making corrections and commenting on areas of improvement for this document. Lastly, I would like to thank my sister, my parents, and my friends for always giving me their support when I needed it.
\end{acknowledgments}

\tableofcontents
\listoffigures

\mainmatter


%
%
%
%
%

%
\chapter{Introduction and Background}

Scone \cite{knowledgenuggets, fahlman2006marker} is an open-source knowledge base system (KBS) designed to store a large collection of knowledge, including both general, common-sense knowledge and domain-specific knowledge. By using multiple inheritance and virtual copy semantics, Scone's knowledge base contains a significant amount of implicit knowledge, much more than what is explicitly defined in the knowledge network. Scone is also equipped with fast inference capabilities implemented with parallel marker-passing algorithms that can answer basic queries about the stored knowledge \cite{fahlman2006marker}. Scone is implemented in Common Lisp and has been under development in Carnegie Mellon University's (CMU's) Language Technologies Institute since around 2003. A tutorial book on the design and usage of Scone is in the process of being written. Until it is published, more information about Scone can be found in Fahlman's Knowledge Nuggets blog \cite{knowledgenuggets}.

By itself, Scone is not an intelligent agent that can make decisions, but acts more as a ``smart memory" that can be used to inform other decision-making processes. One of our long term goals is to give Scone the ability to reason robustly about plans and actions that can be used to guide an intelligent agent. The ERIS (Episodic Reasoning in Scone) subsystem of Scone provides a starting point for \replaced{representing}{how} episodic knowledge \replaced{(actions, events, sequences, and plans)}{can be represented} in Scone to make basic reasoning about it possible \cite{fahlman2014eris}. ERIS introduces an event type to Scone representing knowledge about how a state can change, along with some other machinery to represent concepts like before and after or cause and effect of events.

A production rule system consists of a set of production rules along with a working memory that the rules can access. Each production rule has a left-hand-side precondition that tries to match with the working memory and a right-hand-side action that makes modifications to the memory\added{ or performs some other action}. Production systems have been used as planning systems in cognitive architectures such as Soar \cite{lehman2006gentle, laird2019soar, newell1992soar} and ACT-R \cite{anderson1996act, alma991001570449704436}. However, these systems use production systems for the full range of planning tasks, whereas we plan for different systems to handle\added{ simple,} lower-level\added{ inference} and higher-level planning.

Our main contribution is \textit{Score}, short for \textit{Scone Rule Engine}, that uses production rules for automatic planning and inference. This \textit{rule-based planner} will handle checking and firing for lower-level inference tasks that are not handled by Scone's other systems. We also provide an overview of, but do not implement, a \textit{recipe-based planner} that handles more complex and higher-level planning based on breaking down a goal into subgoals and considering alternative plans. We do this to clarify that the goal of the rule engine is to handle just simple automatic inference\added{s}, not all kinds of complex planning as is the case in some other production systems.

As an example of what capabilities we want our rule engine to have, suppose we are using Score to plan a trip, which can be represented as an event type. If we know that the vehicle used in the trip is an airplane, then the trip involves flying, whereas if the vehicle used is a car, then the trip involves driving. These inferences can be phrased informally as rules saying ``if there is a trip event and the vehicle used is an airplane, then the trip is a flying event" and similarly for the other inference. Our rule engine allows defining these rules and performing these inferences to update the knowledge base automatically.

Before outlining the details of our contribution, we first provide the necessary background knowledge for how Scone itself is structured.

\section{Scone}

\subsection{Elements}

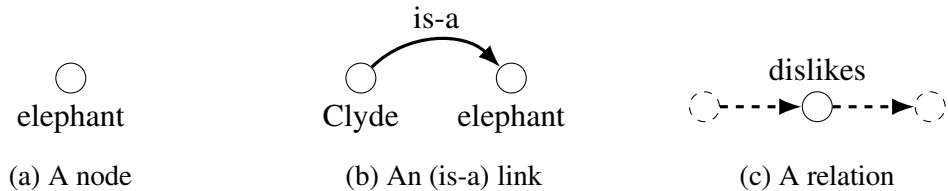
\begin{figure}
    \centering
    \begin{subfigure}[b]{0.30\textwidth}
        \centering
        \begin{tikzpicture}
            \node[draw,circle,label=below:elephant]{};
        \end{tikzpicture}
        \caption{A node}
    \end{subfigure}
    \begin{subfigure}[b]{0.30\textwidth}
        \centering
        \begin{tikzpicture}
            \node[draw,circle,label=below:Clyde] (a) at (0,0) {};
            \node[draw,circle,label=below:elephant] (b) at (2,0) {};
            \draw[is-a-arrow] (a) to[out=45] node[midway, above] {is-a} (b);
        \end{tikzpicture}
        \caption{An (is-a) link}
    \end{subfigure}
    \begin{subfigure}[b]{0.30\textwidth}
        \centering
        \begin{tikzpicture}
            \node[draw,circle,dashed,label=below:\quad] (a) at (0,0) {};
            \node[draw,circle,label=above:dislikes] (b) at (1.5,0) {};
            \node[draw,circle,dashed] (c) at (3,0) {};
            \draw[x-y-z-arrow] (a) to (b);
            \draw[x-y-z-arrow] (b) to (c);
        \end{tikzpicture}
        \caption{A relation}
    \end{subfigure}
    \caption{Scone elements visualized}
    \label{fig:scone_elements}
\end{figure}

The basic unit of knowledge in Scone is an \textit{element}, which is represented as a Common Lisp data structure. Scone elements are denoted using curly braces. Elements are divided into three categories: \textit{nodes}, \textit{links}, and \textit{relations}, as shown in Figure \ref{fig:scone_elements}. A \textit{node} represents a description of some conceptual knowledge about the world, like \{elephant\} or \{the mother of Clyde\}. A \textit{link} describes the relationship between different elements, like \{Clyde is an elephant\} or \{I dislike brussels sprouts\}. There are many different kinds of links that say different things about the elements they are connected to. A \textit{relation} represents some template relationship that can be instantiated to form special links called statement links. For example, \{dislikes\} is a relation, and an instantiation \{I dislike brussels sprouts\} is a statement link.

A specific kind of node in Scone is the \textit{role node}. A \textit{role node} is attached to another node, called the owner node, and represents some concept that the owner node may have or possess. For example, a person may have a mother and a dish may have ingredients, so we can have in Scone that \{mother\} is a role node with owner \{person\} and \{ingredient\} is a role node with owner \{dish\}.

Nodes are divided into \textit{type nodes} and \textit{individual nodes}. A \textit{type node} represents a typical member of some set, like \{elephant\} represents a typical elephant. An \textit{individual node} represents a specific member of some set, like \{Clyde the elephant\} represents a specific elephant named Clyde. These notions also extend to roles. A \textit{type role} represents a typical role that the owner may possess multiple of, like \{ingredient\} of a dish. An \textit{individual role} represents a specific role that the owner generally only possesses one of, like \{mother\} of a person.

An \textit{intersection type} is a type node that is the intersection of several different types. For example, the type \{man\} can be seen as the intersection of the types \{male\}\added{, \{human\},} and \{adult\}. A \textit{defined type} is an intersection type that additionally has a predicate that must be satisfied for nodes to be an instance of that type. As we explore later, we see that intersection types can be seen as simple if-then rules: if an element is a \{male\}\added{, a \{human\},} and an \{adult\}, then the element is a \{man\}.

Each link has a set of \textit{wires} that control how links connect different elements together. Two important wires are the A-wire and B-wire of a link. The A-wire is connected to the first element referenced in the link, and the B-wire is connected to the second element referenced in the link. For example, in the link representing \{Clyde is an elephant\}, the A-wire is connected to \{Clyde\} and the B-wire is connected to \{elephant\}.

This last example is an instance of a special kind of link called an \textit{is-a link}. These links state that the element connected to the A-wire is a more specific concept of the element connected to the B-wire. For example, an is-a link connecting \{human\} to \{mammal\} represents the knowledge that a human is a specific kind of mammal. There are also \textit{eq links} that state the elements connected to the A-wire and B-wire are equivalent. The set of all is-a and eq links forms what is called the is-a hierarchy. Going ``up" the is-a hierarchy means going from the A-wire element to the B-wire element of is-a links which leads to more general concepts, and going ``down" the is-a hierarchy means going from the B-wire element to the A-wire element of is-a links which leads to more specific concepts. Elements above a given element in the is-a hierarchy are called its \textit{superiors}, and elements below a given element are called its \textit{inferiors}. Superiors that are type nodes are called \textit{supertypes}, and inferiors that are type nodes are called \textit{subtypes}. (All superiors are in fact supertypes because individual nodes cannot have any inferiors.) The topmost element in the is-a hierarchy is a type called \{thing\}: all other concepts are more specific than \{thing\}.

\subsection{Multiple Inheritance and Marker Passing}

Scone supports \textit{multiple inheritance} through its is-a hierarchy: each element can have any number of incoming and outgoing is-a links. Inheritance allows Scone to infer many facts about the knowledge base that are not explicitly stated. For example, if the knowledge base contains knowledge that birds are feathered and that a chicken is a type of bird, the fact that a chicken is feathered is inherited from the bird type and can be inferred.

Scone also has cancel-links to support reasoning with exceptions. For example, we can have in the knowledge base that \{bird\} is a subtype of \{flying thing\}, since birds generally can fly, and that \{penguin\} is a subtype of \{bird\}. However, since penguins can't fly, we add a cancel-link from \{penguin\} to \{flying thing\} to indicate that we do \textit{not} want to inherit \{flying thing\} from \{bird\}. Cancelling is a complex topic that can occasionally lead to knowledge ambiguities, and we do not deal with them extensively in the rule engine.

Scone uses a marker-passing system to perform efficient inference and to handle inheritance and virtual copying \cite{fahlman2006marker}. Each node in Scone is equipped with a fixed set of marker bits that can be turned on and off. These markers can be conditionally turned on and off in parallel. For example, if a set of nodes is marked with some marker $m$, we can look at all is-a links in parallel and request for each one to mark the node attached to the B-wire with $m$ if the node attached to the A-wire is marked with $m$. This operation takes all nodes marked with $m$ and marks with $m$ all nodes one level above them in the is-a hierarchy. By repeating this operation until no new nodes are marked with $m$, we can quickly find all nodes above a specific node in the is-a hierarchy, and this procedure is called an \textit{upscan}. There is a corresponding procedure called \textit{downscan} where we mark all nodes below a specific node in the is-a hierarchy.

\subsection{Roles and Virtual Copy Semantics}

Role nodes give rise to special links called \textit{has links}. When a new role is declared for an owner, a new \textit{has link} is also created with A-wire connected to the role node and B-wire connected to the owner node. This link signifies that the owner type possesses some element of the role type. Any inferior of the role is called a player or a filler, indicating that ``the player is a role filler of the owner."

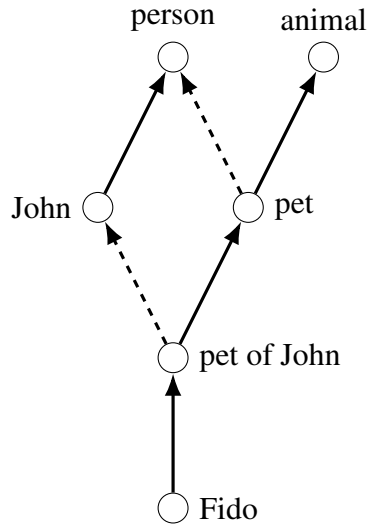
\begin{figure}
    \centering
    \begin{tikzpicture}
        \node[draw,circle,label=above:person] (person) at (0,6) {};
        \node[draw,circle,label=above:animal] (animal) at (2,6) {};
        \node[draw,circle,label=left:John] (John) at (-1,4) {};
        \node[draw,circle,label=right:pet] (pet) at (1,4) {};
        \node[draw,circle,label=right:pet of John] (petj) at (0,2) {};
        \node[draw,circle,label=right:Fido] (Fido) at (0,0) {};
        \draw[is-a-arrow] (John) -- (person);
        \draw[x-y-z-arrow] (pet) -- (person);
        \draw[is-a-arrow] (pet) -- (animal);
        \draw[is-a-arrow] (petj) -- (pet);
        \draw[x-y-z-arrow] (petj) -- (John);
        \draw[is-a-arrow] (Fido) -- (petj);
    \end{tikzpicture}
    \caption{Role nodes visualized for the knowledge ``Fido is a pet of John." Here \{pet of John\} is a virtual copy of \{pet\} that \{John\} inherits from \{person\}.}
    \label{fig:role_intro}
\end{figure}

Scone implements \textit{virtual copy semantics} to handle complex knowledge about roles in a consistent way \cite{fahlman2006marker}. When a new inferior of a type with several roles is created, the roles are virtually copied through inheritance. This means that the new inferior is treated as if it has a copy of each of the roles of its parent type, though no actual copying is done. For example, suppose a \{pet\} role is defined with owner \{person\}. This creates a has-link \{people have pets\}. An individual person \{John\} then inherits the has link and the \{pet\} role, creating a virtual copy \{pet of John\}. To say that an individual \{Fido\} is a pet of John, we then simply add an is-a link from \{Fido\} to \{pet of John\}. This is visualized in Figure \ref{fig:role_intro}, with solid arrows denoting is-a links and dashed arrows denoting has links.

Scone also contains some marker operations related to role nodes and relations. For role nodes, there is a function that marks every node $x$ such that $x$ is a ROLE of OWNER for a specified ROLE and OWNER, and a function that marks every node $x$ such that PLAYER is a ROLE of $x$ for a specified ROLE and PLAYER. For relations, there are functions that marks every node $x$ such that ``$x$ RELATION B" or ``B RELATION $x$" is true when RELATION and either A or B are specified.

Using these marker operations, we can efficiently answer a variety of queries about the knowledge base, such as if Clyde is an animal or if Clyde has any siblings. These marker passing operations will form the basis for checking if knowledge in Scone satisfies any production rules that can potentially fire. Virtual inheritance will also play a role in rule ``trigger" activation.

\subsection{Multiple Contexts and Episodic Reasoning}

All knowledge in the Scone knowledge base exists under some \textit{context}, which can be any Scone element. By default, all knowledge exists in a large context called \{general\}. Scone implements a \textit{multiple-context mechanism} that can handle different knowledge existing in different contexts and switching between contexts. Each element in Scone has a \textit{context-wire} connected to a context that denotes when the knowledge represented by that element exists or is valid.

To activate a context, a special context marker is placed on the context element, then upscanned to all supertypes of the context. This allows Scone to easily consider alternative ``universes" that are mostly the same as our universe but with some differences. For example, suppose we want to model knowledge about the Harry Potter universe. We can create a \{Harry Potter universe\} context that is a subtype of the \{general\} context. When activating the \{Harry Potter universe\} context, we inherit all the knowledge in \{general\}, so many basic facts such as ``humans have two hands" are still true in this context. However, we can add nodes and links to just this context, such as \{wizards can use magic\}, that are only true when the \{Harry Potter universe\} context is active. When we reactivate the \{general\} context, the context-wire of the new nodes and links refer to the \{Harry Potter universe\} which no longer has the active context marker, so they are inactive. Scone's multiple-context mechanism allows easy exploration of different possible states containing slightly different knowledge.

Episodic reasoning in Scone \cite{fahlman2014eris} depends heavily on the multiple-context mechanism. Scone has an \{event\} type that represents an event with a ``before" state and an ``after" state. We represent this by giving \{event\} two roles \{before-context\} and \{after-context\}. As their names imply, \{before-context\} is a context containing knowledge that is true before the event occurs, and \{after-context\} correspondingly contains knowledge that is true after the event occurs. Generally, the \{before-context\} is inherited from the active context, meaning all the currently active knowledge is valid before the event, and the \{after-context\} inherits from \{before-context\} and makes a few modifications. As an example, suppose we have an event representing a caterpillar undergoing metamorphosis into a butterfly. In the before-context, we would have an is-a link from an individual to \{caterpillar\}, and in the after-context, we would cancel this link and have a new is-a link from the individual to \{butterfly\}. Scone also has an \{action\} type, which is just a special \{event\} that has an \{agent\} role representing something that \textit{causes} the event. An agent could be a \{person\} in an event like throwing a ball.

\section{Thesis Structure}

The remainder of the thesis is structured as follows. In Chapter \ref{related}, we discuss related work in production systems as they are used in cognitive architectures. In Chapter \ref{design}, we design a structure for production rules that is compatible with Scone, and we also describe the recipe-based planner to provide context for our design goals. Chapter \ref{checking} contains a description of rule triggers and a rule search algorithm that are used to check and fire these rules efficiently with the knowledge base. Finally, we conclude with future directions for research in Chapter \ref{conclusion}. Some concrete code and a pointer to the rule engine codebase is provided in the Appendix.

\chapter{Related Work} \label{related}

\section{Rule-Based Systems}

Production rule systems have been used in existing cognitive architectures such as Soar and ACT-R. In general, a production system consists of a set of rules (or productions), which can be thought of as if-then statements, and a mechanism for checking and firing the rules. Production systems assume some form of memory or knowledge that the productions can read and modify. The ``if" part of a production defines some condition that must be matched by the current state of the memory. If it is matched, the ``then" part of that production is fired and \replaced{executes some action, which often includes modifying}{modifies} the current memory or knowledge. After a production is fired, the memory can be modified to a state that matches the ``if" part of a different rule, causing a chain of production rule firings.

This general structure leaves many parts of the system up to the underlying architecture to design and implement, such as how the memory is stored and matched and what to do if multiple production rules match the memory and can fire. Both Soar and ACT-R use this same general structure for their production systems, though they address design questions such as these differently and emphasize somewhat different parts of their systems.

\subsection{Soar}

Soar was first developed by Allen Newell's research group in the CMU Department of Computer Science, and work on Soar \replaced{has continued}{continues} primarily under John Laird at the University of Michigan and Paul Rosenbloom at the University of Southern California \cite{lehman2006gentle, laird2019soar, newell1992soar}. It is designed to be a unified cognitive architecture, a theory that can explain how cognition works across a wide range of tasks. Some of its key design objectives include being goal-oriented, meaning it makes rational decisions in order to reach a defined goal, and requiring use of large amounts of knowledge and abstractions, which informs its decisions given what it currently perceives about the state.

Soar distinguishes between two kinds of memory: long-term memory (LTM) and working memory (WM). Long-term memory is further divided into procedural, semantic, and episodic knowledge, though procedural LTM is the main kind of knowledge used in the rule system. Procedural knowledge stores information about how and when to carry out tasks to reach a goal and is phrased in terms of if-then production rules. Working memory contains information about the current state, such as the goal and any relevant information about the environment. Soar additionally differentiates between \textit{rules} and \textit{operators}: rules can \textit{augment} the working memory state and make suggestions about what operator to choose, whereas only operators can \textit{modify} knowledge stored in working memory.

LTM and WM work together to achieve a desired goal through a sequence of five phases: input, elaboration, decision, application, and output. The input phase is when the goal is defined and knowledge about the starting state is brought into WM from a perception module. The elaboration phase is when production rules in procedural LTM fire. Since these rule actions are monotonic, meaning they only augment the current WM with new information and suggestions about what possible operator to take, all applicable rules for the current state fire in parallel. The elaboration phase ends when no more rules can be fired.

After the elaboration phase ends and the decision phase begins, the system considers all of its suggested operators and chooses one to apply. The system makes a decision based on some specification of preferences, such as preferring one operator over another or assigning each operator a preference score and choosing the operator with the highest score. After choosing an operator, the system applies it to modify the working state in the application phase. The output phase consists simply of sending the operator to an output module like one that controls motor output.

\subsection{ACT-R}

ACT-R, short for Adaptive Control of Thought-Rational, is developed primarily by John Anderson's research group in the CMU Department of Psychology \cite{anderson1996act,alma991001570449704436}. It is another cognitive architecture that aims to model and explain the mental processes that are central to human cognition. The key assumption in ACT-R is that there are two kinds of knowledge: declarative knowledge and procedural knowledge. Declarative knowledge is represented as ``chunks," structures with an isa pointer to what kind of fact it is and additional pointers to content in the fact. Procedural knowledge is represented as production rules in the form of if-then statements.

In ACT-R, many different modules are responsible for processing different kinds of information, such as a visual module for processing visual information and a declarative module for retrieving declarative knowledge. Within each module, processing can be done in parallel, and outputs are written to a small buffer. Writing data to and reading data from the buffers is serial, creating communication bottlenecks between different modules. The modules are connected by a central procedural module, which coordinates data from the different buffers and processes them with production rules. The procedural module is designed to replicate cognitive mechanisms linked to the basal ganglia in the brain, which is hypothesized to take in information from disparate regions of the brain and process them to make overarching decisions.

The production system in the procedural module is designed so that only a single production rule can fire at a time, creating a central bottleneck where there must be serial processing of information at the procedural level. One reason to have this bottleneck is to prevent multiple rules firing and causing contradictory changes to the knowledge; another is that there is evidence from cognitive psychology suggesting a central bottleneck in human brains when processing a problem state \cite{borst}. When multiple rules are applicable, the system chooses a rule with the highest utility for some utility function. The procedural module completes tasks by reading in information and applying production rules until the goal is reached. Researchers have conducted experiments showing that human cognitive processing in a variety of tasks can be replicated by suitably designed ACT-R modules.

There has also been some research on integrating Scone with ACT-R \cite{oltramari2012pursuing}. In this research, Oltramari and Lebiere add Scone as a knowledge module to ACT-R, creating ACT-RK (where the K stands for Knowledge). The unified ACT-RK has improved knowledge-based reasoning, as was demonstrated when it was applied to the task of semantically describing visual input in the form of video.

\subsection{Other Related Work}

Some rule systems, such as Prolog-based rule engines \cite{bratko2001prolog}, use backward chaining instead of forward chaining. In Prolog systems, programs are built from facts and rules saying that a conclusion is true if some premises are true. Execution is driven by a query, such as \texttt{?- animal(X).} asking what things are animals. Such a query causes backtracking among the rules to find a resolution for the query, in this case outputting everything that can be deduced to be an animal from the rules.

In systems with a large amount of knowledge and many rules, naive rule pattern matching is often too slow. One optimization is the Rete algorithm \cite{forgy1989rete} that can improve matching performance. The Rete algorithm builds a network of nodes containing tests for knowledge that partially satisfies rule predicates. A path from a root to a leaf indicates an entire rule predicate that is satisfied. When new facts are added, they are propagated down the network, and any leaf nodes that are reached indicate that the action is ready to fire.

The Scone system manages and stores semantic knowledge, though it is not the only semantic knowledge network. The Semantic Web \cite{herman_2013} is an extension to the World Wide Web that enables semantic analysis of data from web pages. It is comprised of several layers, including RDF (Resource Description Framework) and OWL (Web Ontology Language). RDF makes statements about data using subject-predicate-object triples. A collection of triples forms a network of information about web resources. OWL is based on formal logic and consists of property assertions about what kinds of relationships between different terms are allowed. The structure of OWL can be exploited by computer programs to verify logical consistency and to make implicit knowledge explicit through deduction.

\chapter{Scone Planning Systems} \label{design}

The long-term goal is to have two planning systems integrated with Scone: one ``rule-based" planner and one ``recipe-based" planner. The goal of the rule-based planner is to use productions to perform fast automatic thinking, and the goal of the recipe-based planner is to be slower and more deliberative in considering how to carry out the steps in a ``recipe plan." Both kinds of planners are valuable and complement each other: the rule-based planner can answer a large range of simple queries and automatically augment its knowledge, and the recipe-based planner can leverage this knowledge to create more complex plans to achieve its goals.

In this chapter, our main contribution is a design for production rules that interfaces properly with the knowledge structure of Scone. We have implemented this design in Common Lisp code, allowing the user to define new rules that can be added to the knowledge base. Implementation of the recipe-based planner is beyond the scope of this research and left to future work; we describe how it should behave to illustrate that the rule-based planner is not designed to carry out complex planning.

Scone stores a rich representation of a large amount of knowledge and makes that knowledge simple to query for other applications built on top of it. In that sense, Scone has much in common with the memory modules in Soar and ACT-R. More specifically, the Scone architecture with its element structure can be seen as analogous to declarative knowledge in ACT-R, and our new production system corresponds to procedural knowledge in ACT-R. Scone can also be compared to working memory in Soar, though much larger in scope as working memory is generally quite small, and the new production system plays a similar role as Soar's procedural long term memory.

One important distinction between the rule-based planner in Scone and the production systems in Soar and ACT-R is that Soar and ACT-R depend on production rules to carry out complex planning with goals. We expect the recipe-based planner to handle goal-based planning, not the rule-based planner. Correspondingly, our productions are not designed to work towards specific goal states in mind but are instead meant to represent fast and reflexive thinking.

\section{Production Rule Design}

The rule-based planner relies on production rules to perform fast and reflexive thinking. To this end, we introduce a new kind of knowledge to the Scone knowledge base system called a \textit{rule} (short for production rule). Scone rules are similar to ordinary production rules in that they have an ``if" precondition, also known as the left-hand-side (LHS) of the rule, and a ``then" action, also known as the right-hand-side (RHS). However, the analogous ``working memory" would be the entire Scone knowledge base, which is quite different from the working memories of other production systems because it is much larger, so the production rules must be structured to work nicely with Scone and its various mechanisms.

The ``if" part of a rule defines some predicates that must match with some portion of the knowledge base in order to satisfy the rule. Matching in Scone is a little more complicated than simple pattern matching because inferiors in the is-a hierarchy should also be considered when matching. Since the Scone knowledge base is generally very large, we only want to consider a localized portion of the knowledge base in a single rule. The most degenerate form of a rule only considers a single element and specifies some predicates about it. For example, one possible rule is that if an element is a \{male\} as well as an \{adult\}, then that element is a \{man\}. These rules that only depend on a single element are in fact examples of Scone's existing defined types or intersection types, so defined and intersection types can be seen as special cases of rules.

For rules that consider multiple elements in the LHS, the elements should be connected to each other in some way through roles and/or relations. For example, a rule could depend on a meeting and the start time of that meeting, which would be two elements with one of them being a role of another. Conceptually, a rule should say something about a local set of elements, so a rule that depends on multiple elements that are not directly connected to each other through roles or relations makes little sense.

The action of a rule defines what to do if the rule LHS is satisfied by some set of elements. An action can be something like adding links between some of the elements in the rule. These modifications to the knowledge base would fall under the active context. An action could also cause external actions such as motor movements if Scone is connected to a motor module. The result of an action could cause the knowledge base to end up in a state that satisfies other rules, leading to forward chaining of rules that each fire consecutively.

\subsection{If-added and if-needed rules}

We separate Scone rules into two different classes: \textit{if-added} rules and \textit{if-needed} rules. \textit{If-added} rules, also called \textit{eager} rules, are fired when new links are added to the knowledge base. \textit{If-needed} rules, or \textit{lazy} rules, are fired when some value in the knowledge base is requested but not found in the knowledge base, and they compute the requested value and save it in the knowledge base as their action.

We distinguish between these two kinds of rules so that they can serve slightly different purposes. A rule should be an if-added (eager) rule if the action of the rule should be executed immediately after the LHS is satisfied, whether because it could cause other rules to fire or because the action is urgent and needs immediate attention. If-added rules are more similar to the production rules seen in existing production systems. They check if the memory, which in Scone's case is the entire knowledge base, matches their predicates, and they fire their actions if so. These actions can then change the memory, causing other rules to match and fire. We only check if-added rules when new knowledge is added (hence their name of ``if-added" rules), since if the knowledge remains the same, the set of rules that are satisfied remains the same. Only the addition of new knowledge can cause rules to fire. When this happens, the system eagerly fires rules until it reaches a point where no further rules can fire, at which point it pauses until more knowledge is added in the future. The trade-off we are making with eager rules is that adding new knowledge takes a little more time than before, but retrieving knowledge from the knowledge base remains fast with no additional overhead.

Contrasting with if-added rules, a rule should be an if-needed (lazy) rule if a knowledge slot is empty and can be computed but computing it is not a high priority. These rules defer rule checking to when some knowledge is explicitly requested and not found. When this happens, the system checks to see if there is suitable knowledge in the knowledge base that can match the predicates, and uses it to compute the requested knowledge if so. After executing the rule action that does the computation, the rule engine adds the new knowledge to the knowledge base, which can be reused to avoid further rule firing if the same knowledge is requested again in the future.

\section{Rule Structure}
We now give a detailed outline of how rules are structured in Scone and provide some examples. Rules consist of several parts:
\begin{itemize}
    \item The \textit{variables} of a rule create placeholders where Scone elements can be plugged into the rule. This creates an abstraction where the rule can be applied to any valid elements, not just to some specific individuals. For each variable, a Scone \textit{type constraint} can be specified such that an element must be an inferior of the specified type to satisfy the rule. Each variable can have at most one specified type constraint, though specifying an intersection type effectively allows any number of type constraints.
    
    Some rule variables can also optionally be tagged as \textit{proper}, signifying that only proper Scone nodes can be substituted in for the variable (i.e. no generic role nodes). We will see why this is needed later.

    \item The \textit{x-y-z-predicates} of a rule define the relationships between different elements in the rule. Each predicate consists of three elements in the form (X Y Z). The Y element here must be a role or relation, and the X and Z elements are variables or Scone elements. If Y is a role, the predicate represents ``X is a Y of Z." If Y is a relation, the predicate represents ``statement X Y Z is true." Each predicate must be true for the rule to be satisfied.

    \item The \textit{action} of a rule is executed when the rule is fired and the predicates of the rule are satisfied for some elements in the knowledge base. The possible actions of a rule are slightly different for if-added and if-needed rules.
    \begin{itemize}
        \item For if-added rules, the action can be arbitrary code that uses the variables defined in the rule. The rule will substitute the satisfying elements into the variables and execute the code, causing either changes in the knowledge base or performing some action in the real world.
        \item For if-needed rules, the action takes the form (X Y Z), where X is an arbitrary computation, Y is an individual role node, and Z is a variable representing the owner of the desired value. The action is executed by carrying out computation X and setting it as the desired slot filler ``the Y of Z."
    \end{itemize}
\end{itemize}

The variables with their type constraints together with the x-y-z-predicates form the LHS predicates of the rule, and the action is the RHS of the rule. We additionally place a constraint on the x-y-z-predicates of a rule to make sure that the elements in a rule are connected to each other. If we consider the graph with the X and Z values as the vertices and an edge between X and Z for each x-y-z-predicate, this graph must be connected for the rule to be well-defined. For example, a rule with four variables A, B, C, and D with only x-y-z-predicates ``A is the mother of B" and ``C is the mother of D" is ill-defined because variables A and B are not connected to variables C and D. If we allowed such rules, then any time we create a mother link, the new mother link taken together with any other existing mother link would satisfy the rule, causing an overabundance of rule firings. These kinds of rules also make little conceptual sense, since elements in a rule should be related to each other in some way.

We describe the following example rules using language to convey what the rules mean conceptually. Code in Common Lisp to define the rules can be found in the Appendix.

\subsection{If-added rule example} \label{if-added-example}

Consider an example if-added rule: if someone is traveling and the vehicle they are in is an airplane, then that person is flying. We represent \{traveling\} as an event type with a \{travel vehicle\} role. The parts of this example rule would be as follows:

\begin{itemize}
    \item There are two \textit{variables} A and B that represent the travel event and the vehicle, and variable B has the \textit{type constraint} ``B is an airplane".
    \item There is one \textit{x-y-z-predicate} ``B is the travel vehicle of A" that captures the relationship between A and B.
    \item The \textit{action} adds the link ``A is a flying event" to the knowledge base.
\end{itemize}

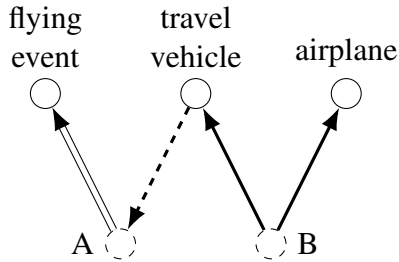
\begin{figure}
    \centering
    \begin{tikzpicture}
        \node[draw,circle,label=left:A,dashed] (A) at (1,0) {};
        \node[draw,circle,label=right:B,dashed] (B) at (3,0) {};
        \node[draw,circle,label={[align=center]above:travel\\vehicle}] (travel vehicle) at (2,2) {};
        \node[draw,circle,label=above:airplane] (airplane) at (4,2) {};
        \node[draw,circle,label={[align=center]above:flying\\event}] (flying event) at (0,2) {};
        \draw[is-a-arrow] (B) -- (travel vehicle);
        \draw[x-y-z-arrow] (travel vehicle) -- (A);
        \draw[is-a-arrow] (B) -- (airplane);
        \draw[res-is-a-arrow] (A) -- (flying event);
    \end{tikzpicture}
    \caption{A diagram of the if-added flying event rule with some of the relevant elements. A and B are drawn with dashed circles to indicate that they are placeholders for other elements. Solid lines indicate is-a links, and dashed lines indicate x-y-z links (in this case a role has-link). The double lines indicate links that are created when the rule action is fired.}
    \label{fig:reading_rule}
\end{figure}

The portion of the knowledge base that would match with this rule is visualized in Figure \ref{fig:reading_rule}. Once this if-added rule is defined and put into the system, the rule engine will start listening to additions to the knowledge base. When it detects new knowledge that could potentially match certain elements to A and B in the predicates, it starts trying to check the rule and executes the action if the predicates are satisfied.

One part of the rule to note is that we want to specify a type constraint for variable B but not for variable A. The reason for this is that variables should only contain a type constraint if the type of that variable is essential to the truth value of the rule predicates. The x-y-z-predicates of a rule place some implicit type constraints on the variables. For the example rule, the fact that A is a \{traveling\} event and B is a \{vehicle\} can be inferred from the x-y-z-predicate. Knowing that a candidate element for B is an airplane is essential for determining if A is a \{flying\} event, since if B was a different vehicle like a car, A would be a different kind of traveling event. On the other hand, there is no need to specify any explicit type constraints for A, since the implicit ones are enough to determine if A is a flying event.

As a side note, the rule engine will still work if extraneous type constraints are specified, just at the cost of potentially more work done. This is because a \textit{trigger} is created for each type constraint that may be checked whenever new knowledge is added. See Section \ref{triggers} for details regarding rule triggers.

\subsection{If-needed rule example} \label{if-needed-example}

Now consider an example if-needed rule: if we want to know the duration of a meeting, and we know that the start time and end time of that meeting are $t_1$ and $t_2$, then we can compute the duration as $t_2 - t_1$. The parts of this rule would be as follows:

\begin{itemize}
    \item There are three \textit{variables} A, B, and C that represent the start time, end time, and the meeting. Variables A and B should be tagged as \textit{proper} to indicate that only proper values should be substituted for them in the rule.
    \item There are two \textit{x-y-z-predicates} ``A is the start time of C" and ``B is the end time of C."
    \item The \textit{action} computes B$-$A using the proper values for A and B and returns it as the answer ``the duration of C."
\end{itemize}

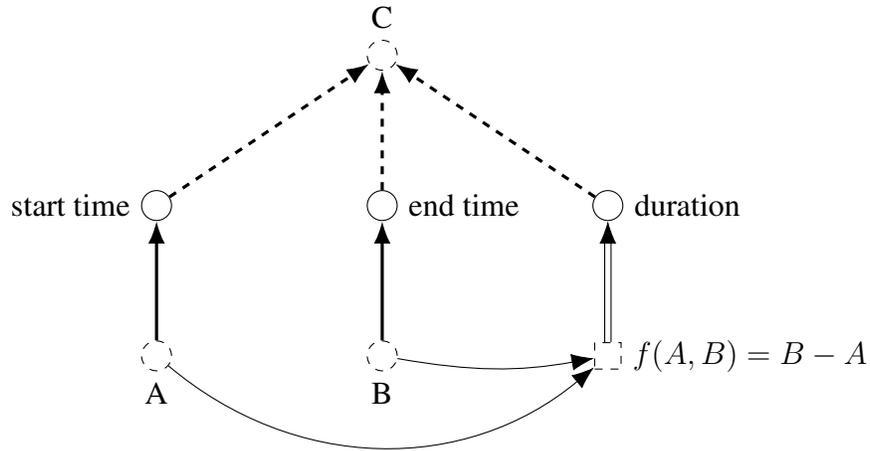
\begin{figure}
    \centering
    \begin{tikzpicture}
        \node[draw,circle,label=below:A,dashed] (A) at (0,0) {};
        \node[draw,circle,label=below:B,dashed] (B) at (3,0) {};
        \node[draw,circle,label=above:C,dashed] (C) at (3,4) {};
        \node[draw,circle,label=left:start time] (st) at (0,2) {};
        \node[draw,circle,label=right:end time] (et) at (3,2) {};
        \node[draw,circle,label=right:duration] (d) at (6,2) {};
        \node[draw, dashed, minimum size=10, label=right:{$f(A,B)=B-A$}] (res) at (6,0) {};
        \draw[is-a-arrow] (A) -- (st);
        \draw[is-a-arrow] (B) -- (et);
        \path[x-y-z-arrow] (st) edge (C);
        \path[x-y-z-arrow] (et) edge (C);
        \path[x-y-z-arrow] (d) edge (C);
        \path[-{Latex[length=3mm]}] (A) edge[bend right=40] (res);
        \path[-{Latex[length=3mm]}] (B) edge[bend right=10] (res);
        \draw[res-is-a-arrow] (res) -- (d);
    \end{tikzpicture}
    \caption{A diagram of the if-needed meeting duration rule. The arrows from A and B to the node labeled $f(A,B)$ indicate that A and B will be used for the computation, and the double line from $f(A,B)$ to ``duration of C" represents the is-a link that will be created and returned.}
    \label{fig:meeting_rule}
\end{figure}

The parts of the rule are visualized in \ref{fig:meeting_rule}. After this if-needed rule is defined, the system remembers it as a possible way to compute the duration of the meeting when requested. If the duration of some meeting is requested, the system first checks to see if a value exists for that meeting, returning it if it does. If it does not, then the rule system will try to match knowledge about the meeting to the rule, and if the matching is successful, will compute the requested duration and add it to the knowledge base.

The reason variables A and B should be tagged as \textit{proper} in this rule is to ensure that only nodes with actual values are substituted in, since the rule action needs the values of A and B. A role node can be created and given a type but not a specific value, for example by saying that the start time of a meeting ends in :30 but the exact time is unknown. This makes the start time of that meeting a concrete element that is a subtype of \{time ending in :30\}, but trying to compute the duration of that meeting would fail because there is no proper value for the start time of the meeting.

\section{Recipe-Based Planner}

The recipe-based planner\added{, of which we have partial prototypes but not yet a full implementation,} is based on Fahlman's BUILD planner \cite{ELLIOTTFAHLMAN19741} adapted to the framework of Scone and episodic knowledge representation described in Eris. BUILD describes a flexible planning system in the setting of moving blocks around on a table. In the setting considered, the table starts out with some configuration of blocks, and the goal is to rearrange the blocks into some goal state without causing any instability. The BUILD planner achieves this by breaking up the goal into subgoals and planning out how to achieve each subgoal using different actions. If when planning it reaches a state in which it cannot continue due to instability, it recursively backtracks to a previous state and tries different actions. In this way, the planner can avoid getting stuck in error states and can generally always find options to make progress.

BUILD's central capability of backtracking to previous states and considering different alternatives can be implemented effectively using Scone's multiple context system. In Scone, information about the current state is stored in the active context. When considering an action to perform, the active context fills the \{before-context\} role of the action, and new information about the resulting state will be stored in the \{after-context\} role of the action. The recipe-based planner can consider what happens after performing this action by activating the \{after-context\} and looking at the resulting state. If this eventually leads to an undesirable state where achieving the subgoal is difficult or impossible, then the planner can simply reactivate the \{before-context\} to revert the consequences of the action and try out other possible actions.

The way Scone stores episodic knowledge as described in Eris also guides the way the recipe-based planner breaks down a goal into subgoals and tries alternative strategies for reaching different goals. Suppose the planner is trying to achieve the goal of going to the Pittsburgh airport. This goal \{go to the Pittsburgh airport\} can be represented as an action type in Scone (i.e. this element is a subtype of \{action\}), and it can further have subtypes such as \{drive to the Pittsburgh airport\} or \{take the bus to the Pittsburgh airport\}.  The split action types in the is-a hierarchy represent qualitatively different plans to take: driving to the airport versus taking the bus each require a particular set of considerations, and the two plans are mutually exclusive.

Action types also have a part-of hierarchy induced by \{part of\} roles that correspond to different subgoals in each plan. For example, the \{driving to the Pittsburgh airport\} action can have parts such as \{take the car keys\}, \{put the car keys in the ignition\}, \{take a left turn at so-and-so street\}, and so on. A different action has different parts and therefore different subgoals, like \{take the bus to the Pittsburgh\} has \{go to the bus stop\} and \{pay the bus fare\} as parts. Each part of an action can have further parts until eventually the action reaches a base action such as some motor movement that is easily carried out. If one of the subgoals cannot be carried out for whatever reason, then the planner tries a different set of subgoals to unblock itself: if turning left is impossible because the road is blocked due to construction, the planner can try to find a different path to drive. Failing that, the planner can decide that the current action is impossible to complete and try a different plan: if the car keys cannot be found, then driving is impossible and a different method of going to the airport is required.

\section{Implementation Status}

The core structure of production rules has been implemented in code and added to the main Scone engine. Scone is developed in Common Lisp, so we extended the engine with Common Lisp code to implement production rules. We represent a production rule as a Lisp structure with components for each of its parts, including variables, predicates, and actions. If-added and if-needed rules share the same underlying data structure. Most importantly, we define two new Lisp macros for creating new rules, one for creating if-added rules and one for creating if-needed rules. These macros are intended to be used similarly to Scone functions such as \texttt{new-type} and \texttt{new-is-a} that add knowledge to the knowledge base. Sample macro calls that define the example rules above are provided in the Appendix.

Implementation of the recipe-based planner is beyond the scope of this research and left to future work; we describe its design to provide context for the goals of the production system planner compared to the goals of the recipe-based planner.

Now that we have a framework for adding new rules to the Scone knowledge base system, we need a system for checking and firing these rules. In the next section, we describe the rule checking engine, which we have fully implemented in the Scone engine.

\chapter{Rule-Checking Engine} \label{checking}

\section{Rule-Checking Components}

Scone is a very large knowledge base, and the rules can reference any subset of elements in the knowledge base, so some care must be taken in how rules are checked to be satisfied. Constantly checking all the rules for if they are satisfied is very wasteful, since rules only become satisfied once new knowledge is added, and then only a few rules (if any) would be satisfied. In addition, trying to match every element naively to rule variables is very inefficient, since there are too many elements to check.

To address the first problem, whenever a rule is created, a set of rule ``triggers" are created and attached to all elements involved in the rule. These triggers are then checked when new knowledge is added for if-added rules and when a role value is requested for if-needed rules. To address the second problem, the rule system carries out a rule-checking search algorithm that leverages Scone's marker-passing operations to search for relevant elements that could satisfy the rule.

\subsection{Rule Triggers} \label{triggers}

If-added rules should only be checked when new knowledge is added to the knowledge base, and if-needed rules should only be checked when a role value is requested. Additionally, we only want to check the subset of rules that could match with the new knowledge or requested role. To achieve this, each newly created rule should create \textit{triggers} that control when to check that particular rule. Three types of triggers are created: one for adding new is-a links, one for adding new role or statement links, and one for accessing a role value. Each trigger is placed on a Scone element using element properties, which are key-value pairs that each element can have. The three types of triggers have slightly different forms to deal with the different ways they are activated.

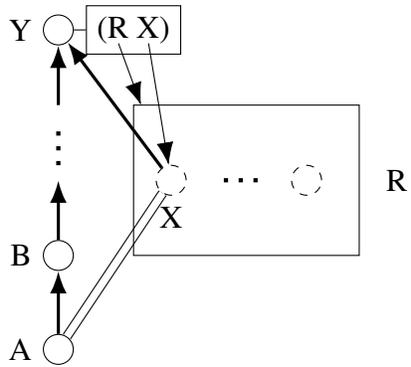
\begin{figure}
    \centering
    \begin{tikzpicture}
        \node[draw,circle,label=left:Y] (Y) at (0,4) {};
        \node[draw] (trigger) at (1,4) {(R X)};
        \node[draw, circle, dashed, label=below:X] (X) at (1.5,2) {};
        \node[draw, circle, dashed] at (3.3,2) {};
        \node[draw, circle, label=left:A] (A) at (0,-0.25) {};
        \node[draw, circle, label=left:B] (B) at (0,1) {};
        \node[] at (4.5, 2) {R};
        \path[-] (Y) edge (trigger);
        \path[-{Latex[length=3mm]}, very thick] (X) edge (Y);
        \path[-{Latex[length=3mm]}, very thick] (A) edge (B);
        \path[-{Latex[length=3mm]}, very thick] (B) edge (0,2);
        \path[loosely dotted, ultra thick] (0,2.2) edge (0,2.7);
        \path[-{Latex[length=3mm]}, very thick] (0,3) edge (Y);
        \path[loosely dotted, ultra thick] (2.2,2) edge (2.7,2);
        \path[-{Latex[length=3mm]}] ([xshift=-2mm,yshift=1.5mm]trigger.south) edge (1.1,3);
        \path[-{Latex[length=3mm]}] ([xshift=2mm,yshift=1.5mm]trigger.south) edge (X);
        \path[-] (A) edge[double, double distance=.1cm] (X);
        \draw (1,1) rectangle (4,3);
    \end{tikzpicture}
    \caption{A visualization of an is-a type constraint trigger. Y is above B in the is-a hierarchy, and has a trigger attached to it with pointers to rule R and variable X. Trigger activation is performed by matching A with X.}
    \label{fig:is-a-trigger}
\end{figure}

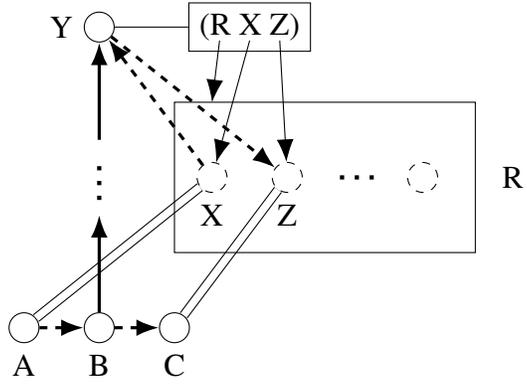
\begin{figure}
    \centering
    \begin{tikzpicture}
        \node[draw,circle,label=left:Y] (Y) at (0,4) {};
        \node[draw] (trigger) at (2,4) {(R X Z)};
        \node[draw, circle, dashed, label=below:X] (X) at (1.5,2) {};
        \node[draw, circle, dashed, label=below:Z] (Z) at (2.5,2) {};
        \node[draw, circle, dashed] at (4.3,2) {};
        \node[draw, circle, label=below:A] (A) at (-1,0) {};
        \node[draw, circle, label=below:B] (B) at (0,0) {};
        \node[draw, circle, label=below:C] (C) at (1,0) {};
        \node[] at (5.5, 2) {R};
        \path[-] (A) edge[double, double distance=.1cm] (X);
        \path[-] (C) edge[double, double distance=.1cm] (Z);
        \path[-] (Y) edge (trigger);
        \path[-{Latex[length=3mm]}, very thick, dashed] (X) edge (Y);
        \path[-{Latex[length=3mm]}, very thick, dashed] (Y) edge (Z);
        \path[-{Latex[length=3mm]}, very thick, dashed] (A) edge (B);
        \path[-{Latex[length=3mm]}, very thick, dashed] (B) edge (C);
        \path[-{Latex[length=3mm]}, very thick] (B) edge (0,1.5);
        \path[loosely dotted, ultra thick] (0,1.7) edge (0,2.2);
        \path[-{Latex[length=3mm]}, very thick] (0,2.5) edge (Y);
        \path[loosely dotted, ultra thick] (3.2,2) edge (3.7,2);
        \path[-{Latex[length=3mm]}] ([xshift=-4mm,yshift=1.5mm] trigger.south) edge (1.5,3);
        \path[-{Latex[length=3mm]}] ([yshift=1.5mm] trigger.south) edge (X);
        \path[-{Latex[length=3mm]}] ([xshift=4mm,yshift=1.5mm] trigger.south) edge (Z);
        \draw (1,1) rectangle (5,3);
    \end{tikzpicture}
    \caption{A visualization of an x-y-z-predicate trigger. Y is above B in the is-a hierarchy, and has a trigger attached to it with pointers to rule R and variables X and Z. Trigger activation is performed by matching A with X and C with Z.}
    \label{fig:x-y-z-trigger}
\end{figure}

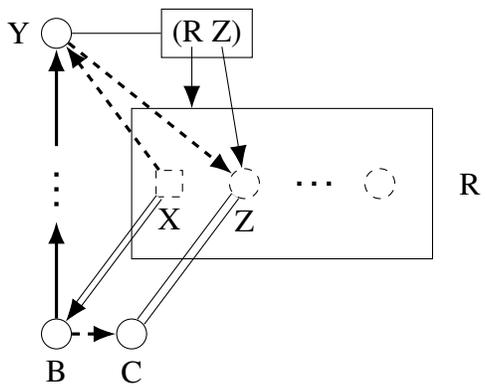
\begin{figure}
    \centering
    \begin{tikzpicture}
        \node[draw,circle,label=left:Y] (Y) at (0,4) {};
        \node[draw] (trigger) at (2,4) {(R Z)};
        \node[draw, dashed, label=below:X, minimum size=10] (X) at (1.5,2) {};
        \node[draw, circle, dashed, label=below:Z] (Z) at (2.5,2) {};
        \node[draw, circle, dashed] at (4.3,2) {};
        \node[draw, circle, label=below:B] (B) at (0,0) {};
        \node[draw, circle, label=below:C] (C) at (1,0) {};
        \node[] at (5.5, 2) {R};
        \path[-] (C) edge[double, double distance=.1cm] (Z);
        \path[-] (Y) edge (trigger);
        \path[-{Latex[length=3mm]}, very thick, dashed] (X) edge (Y);
        \path[-{Latex[length=3mm]}, very thick, dashed] (Y) edge (Z);
        \path[-{Latex[length=3mm]}, very thick, dashed] (B) edge (C);
        \path[-{Latex[length=3mm]}, very thick] (B) edge (0,1.5);
        \path[loosely dotted, ultra thick] (0,1.7) edge (0,2.2);
        \path[-{Latex[length=3mm]}, very thick] (0,2.5) edge (Y);
        \path[loosely dotted, ultra thick] (3.2,2) edge (3.7,2);
        \path[-{Latex[length=3mm]}] ([xshift=-2mm,yshift=1.5mm]trigger.south) edge (1.8,3);
        \path[-{Latex[length=3mm]}] ([xshift=2mm,yshift=1.5mm]trigger.south) edge (Z);
        \draw[res-is-a-arrow] (X) -- (B);
        \draw (1,1) rectangle (5,3);
    \end{tikzpicture}
    \caption{A visualization of an if-needed rule trigger. Y is above B in the is-a hierarchy, and has a trigger attached to it with pointers to rule R and variable Z. Trigger activation is performed by matching C with Z. An is-a link is created between the resulting computation X and ``the B of C" as part of the rule action.}
    \label{fig:x-of-y-trigger}
\end{figure}

\begin{itemize}
    \item Let R be an if-added rule with a variable X that has an is-a type constraint Y. When this rule is defined, a trigger of the form (R X) containing pointers to R and X is created and attached to element Y. Conceptually, this trigger represents the fact that any inferiors of element Y could be substituted for variable X in rule R. All inferiors of Y therefore virtually inherit this rule trigger.

    Suppose a new is-a link or eq link from node A to node B is created. This link could cause any rule attached to a trigger that B virtually inherits to become satisfied. Therefore, we perform an upscan on B to check if there are any triggers (R X) on any superior of B. Each such trigger is activated by substituting in A for variable X in rule R, then starting the rule-checking search algorithm on R. A visualization of this kind of trigger activation is provided in Figure \ref{fig:is-a-trigger}.
    
    \item Let R be an if-added rule with an x-y-z-predicate (X Y Z), where Y is a role or relation and X and Z are variables or nodes. When this rule is defined, a trigger containing pointers to R, X, and Z is created and attached to element Y. This trigger represents the fact that a link (A B C) where B is an inferior of Y could match A with X and C with Z in rule R. All inferiors of Y therefore virtually inherit this rule trigger.

    Suppose a new link (A B C) is created, where B is a role node or a relation. If B is a role, this link looks like ``A is a B of C", and if B is a relation, this link looks like the statement ``A B C". We perform an upscan on B to check if there are any triggers (R X Z) that B should inherit. If X is a Scone element in a trigger, we check if A is an inferior of X, and we do not activate the trigger if not. If X is a variable, we substitute A for X in R and allow activation. We also do this check for Z and C. If activation of this trigger is not blocked and A and C are substituted into their corresponding variables, then the rule-checking search algorithm is then started on R. This matching is visualized in Figure \ref{fig:x-y-z-trigger}.
    
    \item Let R be an if-needed rule with action (X Y Z), where X is a computation, Y is a role, and Z is a variable. When this rule is defined, a trigger of the form (R Z) is created and attached to element Y. This trigger represents that if we want to know the B of C where B is an inferior of Y, then we could find out by activating a rule by substituting C for Z in R. All inferiors of Y therefore virtually inherit this rule trigger.

    Suppose the value ``the B of C" is requested, where B is a role node. We perform an upscan on B to check if there are any triggers (R Z) that B should inherit. Each trigger found is activated by substituting C for variable Z in rule R and starting the rule-checking search algorithm on R. The substitution and resulting link created is visualized in Figure \ref{fig:x-of-y-trigger}.
\end{itemize}

\subsection{Rule Variable Substitution}

One necessary component of rule-checking is the ability to substitute a Scone element for a variable in a rule. A rule R contains data about the variables in the rule, the type constraints on those variables, the x-y-z-predicates of the rule, and the rule action. When the system attempts to substitute an element E for variable X in rule R, it must first validate that E is an inferior of any type constraints on X and that any x-y-z-predicates involving X are satisfied by E. If any of these are false, then the substitution fails and E cannot satisfy the rule when substituted for X. If substitution succeeds, then the rule stores E as the current filler for X. When all variables in a rule are successfully substituted, the rule is satisfied and the action is ready to fire using all the stored elements.

\subsection{Rule-Checking Search Algorithm} \label{search-section}

A rule starts to be checked when at least one element is substituted in for a variable in that rule. This occurs when a trigger for a rule is activated as described previously. The next step is to find a set of Scone elements such that all of them can be substituted in for the rest of the variables while satisfying all of the rule predicates. This is accomplished through a recursive search algorithm that searches through possible elements using Scone marker operations.

At a high level, the algorithm starts by choosing an x-y-z-predicate that has just one variable substituted with an element and one unsubstituted variable. It then uses Scone marker operations to mark each element that can satisfy the predicate when substituting it in for the remaining variable. If the remaining variable is marked as proper, the markers are then restricted to the subset of elements that are proper using another marker operation. For each marked element, the algorithm tries to substitute the element into the rule. If substitution fails, the algorithm proceeds to the next marked element. If there are no more marked elements, the algorithm returns NIL. If substitution succeeds, the algorithm recurses to try to find an element for another variable in the rule. If the algorithm successfully finds elements for all the variables in the rule, it executes the action of the rule using the elements it found. The full algorithm pseudocode can be found in Figure \ref{fig:algorithm}.

\begin{figure}
    \centering
    \begin{mdframed}
    \begin{algorithmic}
    \Procedure{check-rule}{$R$}
    \If {$R$ has an element substituted for each variable}
    \State Add $R$ to a queue to be fired
    \State \Return T
    \EndIf
    \State Choose an x-y-z-predicate $(X\;Y\;Z)$ of $R$ with an element substituted for $X$ or $Z$
    \State Allocate a new marker $m$
    \If {$X$ has an element substituted for it}
    \State Mark with $m$ all elements $E$ such that the predicate $(X\;Y\;E)$ is satisfied
    \EndIf
    \If {$Z$ has an element substituted for it}
    \State Mark with $m$ all elements $E$ such that the predicate $(E\;Y\;Z)$ is satisfied
    \EndIf
    \For{each element $E$ marked with $m$}
    \State $R' \gets $ substitute $E$ for the unsubstituted variable $X$ or $Z$ in $R$
    \If {any predicates in $R'$ are false}
    \State \textbf{continue}
    \EndIf
    \State \textproc{check-rule}($R'$)
    \EndFor
    \State \Return NIL
    \EndProcedure
    \end{algorithmic}
    \end{mdframed}
    \caption{Rule-Checking Algorithm}
    \label{fig:algorithm}
\end{figure}

Note that our constraint that the x-y-z-predicates should form a connected graph ensures this algorithm's correctness in finding all possible sets of elements that could satisfy the rule. The algorithm essentially repeatedly chooses an arbitrary edge connecting a substituted variable with an unsubstituted variable and finds an element that can be substituted. Each of these steps of finding suitable elements is relatively fast, since it only involves a single marker operation starting from a substituted element. Since one marker pair is allocated at each recursion, the recursion depth is limited by the number of marker pairs available. This is not an issue because the recursion depth is equal to the number of x-y-z-predicates which is typically no more than three or four, and the number of marker pairs available is generally around fourteen.

In some cases, adding a piece of knowledge can cause multiple rules to fire. When this happens, we add each satisfied rule to a queue and fire them sequentially. This sequential firing after triggers are checked makes the algorithm comparable to the algorithm used by ACT-R. Rules are fired in order of when their triggers are checked, which is generally starting from triggers attached to elements at the bottom of the hierarchy and working their way up, though this order is not strictly defined. We adopt this method of sequential firing to ensure that rule firing is deterministic and consistent, letting Scone take care of preventing any rule action that would add contradictory knowledge to the knowledge base.

\section{End-to-end Example Rule Firing}

To illustrate how Scone's production system works in practice, we step through Scone's processes for checking and firing an example rule. We use the rule given earlier in Section \ref{if-needed-example} but phrased as an if-added rule instead of an if-needed rule. Note that in some cases such as this one, a rule can be defined as either an if-added rule or as an if-needed rule, and choosing which kind of rule depends on how immediately we want the result of the rule to be updated in the knowledge base. As a reminder, the parts of the rule are as follows:

\begin{itemize}
    \item There are three \textit{variables} A, B, and C that represent the start time, end time, and the meeting.
    \item There are two \textit{x-y-z-predicates} ``A is the start time of C" and ``B is the end time of C."
    \item The \textit{action} adds the link ``B$-$A is the duration of C" to the knowledge base.
\end{itemize}

Defining this if-added rule adds triggers (R A C) to role \{start time\} and (R B C) to role \{end time\} for the x-y-z-predicates. Afterwards, any additions to the knowledge base involving these roles cause the triggers to be checked.

\begin{figure}
    \centering
    \begin{tikzpicture}
        \node[draw,circle,label=left:start time] (Y) at (0,4) {};
        \node[draw] (trigger) at (2,4) {(R A C)};
        \node[draw, circle, dashed, label=below:A] (X) at (1.5,2) {};
        \node[draw, circle, dashed, label=below:C] (Z) at (3.5,2) {};
        \node[draw, circle, dashed, label=below:B] (B) at (2.5,2) {};
        \node[draw, circle, label=below:10:30 AM] (A) at (-1,0) {};
        \node[draw, circle, label=below:meeting 27] (C) at (1,0) {};
        \node[] at (4.5, 2) {R};
        \path[-] (A) edge[double, double distance=.1cm] (X);
        \path[-] (C) edge[double, double distance=.1cm] (Z);
        \path[-] (Y) edge (trigger);
        \path[-{Latex[length=3mm]}, very thick, dashed] (X) edge (Y);
        \path[-{Latex[length=3mm]}, very thick, dashed] (Y) edge (Z);
        \path[-{Latex[length=3mm]}, very thick, dashed] (A) edge (Y);
        \path[-{Latex[length=3mm]}, very thick, dashed] (Y) edge (C);
        \path[-{Latex[length=3mm]}] ([xshift=-4mm,yshift=1.5mm] trigger.south) edge (1.5,3);
        \path[-{Latex[length=3mm]}] ([yshift=1.5mm] trigger.south) edge (X);
        \path[-{Latex[length=3mm]}] ([xshift=4mm,yshift=1.5mm] trigger.south) edge (Z);
        \draw (1,1) rectangle (4,3);
    \end{tikzpicture}
    \begin{tikzpicture}
        \node[draw,circle,label=left:end time] (Y) at (0,4) {};
        \node[draw] (trigger) at (2,4) {(R B C)};
        \node[draw, circle, dashed, label=below:B] (X) at (2.5,2) {};
        \node[draw, circle, dashed, label=below:C] (Z) at (3.5,2) {};
        \node[draw, circle, dashed, label=below:A] (B) at (1.5,2) {};
        \node[draw, circle, label=below:11:30 AM] (A) at (-1,-0.5) {};
        \node[draw, circle, label=below:meeting 27] (C) at (1,-0.5) {};
        \node[] at (4.5, 2) {R};
        \path[-] (A) edge[double, double distance=.1cm] (X);
        \path[-] (C) edge[double, double distance=.1cm] (Z);
        \path[-] (Y) edge (trigger);
        \path[-{Latex[length=3mm]}, very thick, dashed] (X) edge (Y);
        \path[-{Latex[length=3mm]}, very thick, dashed] (Y) edge (Z);
        \path[-{Latex[length=3mm]}, very thick, dashed] (A) edge (Y);
        \path[-{Latex[length=3mm]}, very thick, dashed] (Y) edge (C);
        \path[-{Latex[length=3mm]}] ([xshift=-4mm,yshift=1.5mm] trigger.south) edge (1.5,3);
        \path[-{Latex[length=3mm]}] ([yshift=1.5mm] trigger.south) edge (X);
        \path[-{Latex[length=3mm]}] ([xshift=4mm,yshift=1.5mm] trigger.south) edge (Z);
        \draw (1,1) rectangle (4,3);
    \end{tikzpicture}
    \caption{A visualization of two trigger activations for the meeting example rule, one for each of the x-y-z role predicates.}
    \label{fig:example-trigger}
\end{figure}
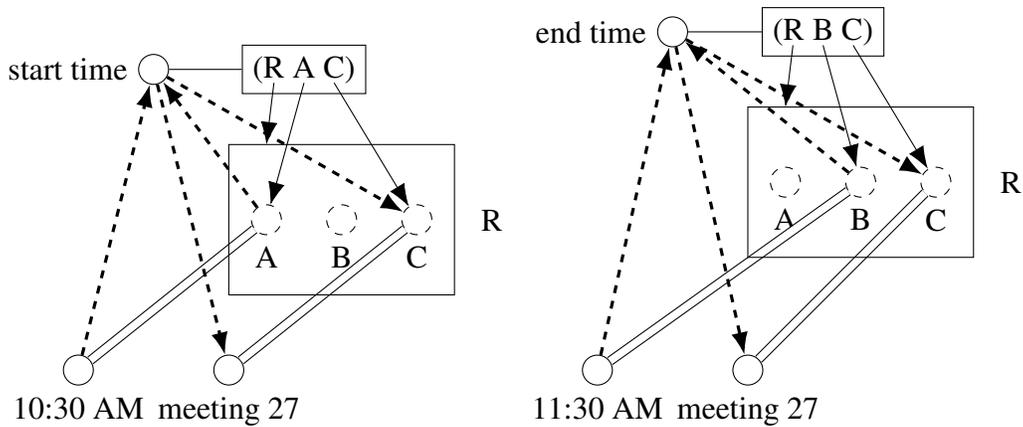

This rule is now defined for an abstract set of elements that would satisfy all the rule preconditions. To see how the rule is specialized to fire for specific individuals, suppose we create a new individual \{meeting 27\} that represents a specific meeting. Now suppose we first add the link ``10:30 AM is the start time of meeting 27" to the knowledge base. This causes an upscan on \{start time\} to find inherited rule triggers that could be activated. There is in fact a trigger (R A C) on \{start time\}, so the rule engine checks if \{10:30 AM\} can be substituted for A and \{meeting 27\} can be substituted for C. There are no explicit type constraints on either A or C and no other violated predicates, so substitution is successful and the trigger is activated, as shown in Figure \ref{fig:example-trigger}.

After this trigger is activated, variables A and C in R are substituted, but variable B is not, so it is the job of the search algorithm to find a suitable element for B. It sees that there is a partially substituted x-y-z-predicate ``B is the end time of meeting 27" in the rule, so it allocates a marker pair and marks all elements E that satisfies ``E is the end time of meeting 27." (For this particular rule, since variable B is marked as proper, the algorithm further restricts the marked elements to only proper elements.) At this point, there are in fact no such elements that can be marked, so the algorithm returns knowing the rule is not satisfied. Conceptually, we currently only know the start time of meeting 27 but not the end time, so as expected we cannot yet compute the duration of the meeting.

Suppose we find out that ``11:30 AM is the end time of meeting 27" and add this to the knowledge base. This causes an upscan on \{end time\} to find inherited rule triggers, which finds trigger (R B C) on \{end time\}. Activating the rule trigger by substituting \{11:30 AM\} for B and \{meeting 27\} for C (as shown in Figure \ref{fig:example-trigger}) is successful because there are no violated predicates, so the search algorithm looks for an element for the unsubstituted variable A.

The algorithm finds the partially substituted x-y-z-predicate ``A is the start time of meeting 27" so it allocates a marker pair and marks all elements E such that ``E is the start time of meeting 27" (again only marking proper elements). This time, it marks \{10:30 AM\} given that ``10:30 AM is the start time of meeting 27" is in the knowledge base. For each marked element E, the rule engine tries substituting E for B in R, so it tries substituting \{10:30 AM\} for B in R. This substitution does not violate any type constraints or x-y-z-predicates, so substitution is successful.

At this point, the rule has elements substituted for all of its variables. This means the rule is satisfied for this particular set of individuals, so the rule action is fired. \{11:30 AM\} $-$ \{10:30 AM\} is computed, yielding \{1 hour\}, and the knowledge ``1 hour is the duration of meeting 27" is added to the knowledge base. This new knowledge could cause rules with triggers that \{duration\} inherits to fire, if any are currently defined.

The process of production rule firing illustrates how Scone can carry out automatic thinking when it gains new knowledge. It only checks rules when it needs to, and it only checks the specific rules that need to be checked, thanks to the rule trigger mechanism. Using the search algorithm, it is able to fetch the relevant knowledge to check if a rule is satisfied.

To elaborate a little more on the choice of making this rule eager or lazy, one reason to make it eager would be if the duration of a meeting has a direct influence on the decision of whether to go to that meeting. For example, suppose we have another if-added rule representing ``if a meeting is at most 2 hours long, then decide to go to the meeting" with the default decision to not go to any meeting over two hours long. We want this additional rule to fire as soon as we know what the duration of a meeting is, and we also want it to fire as soon as we have enough information to find out the duration of a meeting. The first rule provides a way to figure out the duration of a meeting, so defining it as an if-added rule makes sense.

\subsection{If-needed Example}

We could also have reasonably defined the previous rule for computing the duration of a meeting as an if-needed rule. We would want to do so if we wanted the ability to compute this value when someone asks what the duration of \{meeting 27\} is, but do not urgently need the duration of any meeting. The structure of the rule would be essentially identical, but checking the rule would be done differently. Instead of adding two if-added triggers, defining the rule as an if-needed rule adds one trigger (R C) to role node \{duration\}.

If-needed rules are not checked until a value is explicitly requested, so adding knowledge ``10:30 AM is the start time of meeting 27" and ``11:30 AM is the end time of meeting 27" should not activate any triggers. The absence of any inherited triggers on \{start time\} and \{end time\} ensures this behavior. Once the value for the slot filler \{the duration of meeting 27\} is requested, an upscan is performed on role node \{duration\}, which finds trigger (R C). This trigger is activated by substituting \{meeting 27\} for C in R, which is successful because it doesn't violate any predicates.

The search algorithm now looks for elements for variables A and B. Both x-y-z-predicates are partially substituted, so one of them can be arbitrarily chosen, say ``A is the start time of meeting 27." Skipping the details of the marker passing and substitution, the algorithm finds that \{10:30 AM\} can be substituted for A in R. Then, the algorithm looks for an element C that can satisfy ``C is the end time of meeting 27." It finds element \{11:30 AM\} and substitutes it in for C in R.

Now that all variables have been substituted, the rule action ``B$-$A is the duration of B" can be fired. The computation of this rule action returns the value \{1 hour\}. The rule action adds the computed link ``1 hour is the duration of meeting 27" to the knowledge base and returns \{1 hour\} as the requested slot filler. Note that after this rule action executes once, the next time \{the duration of meeting 27\} is requested, the knowledge already exists in the knowledge base, so there is no need to check the rule again.

\chapter{Conclusion and Future Work} \label{conclusion}

In this thesis, we proposed a design for if-added and if-needed production rules that are compatible with the Scone knowledge base system. We also implemented triggers and methods to check efficiently if rules are satisfied. Our work primarily falls under the ``rule-based planner" we have in mind for Scone that performs low-level automatic thinking in response to added or requested knowledge. We provided an outline of the design of the ``recipe-based planner" to provide context for the design goals of both kinds of planners.

Future research into production systems for Scone could extend the capabilities of production rules in various ways. One extension is supporting if-removed rules that would contain negative type constraints in the rule predicates, such as ``A is not an elephant." These rules would fire when links are removed or canceled. Supporting this would involve using Scone's cancel-link capabilities, and ensuring logically consistent behavior would be a challenge. We would also want to apply the rule engine to a much larger knowledge base with a large collection of rules to reveal any potential scaling or consistency issues.

One other possibility for rule checking is \textit{rumination}, a technique used by Learning Reader to perform off-line inferences to learn from natural language \cite{forbus2007integrating}. In the context of Scone, if the system is currently processing a low-priority rule but needs to spend resources on other higher priority tasks, it can put off checking the rule until some point in the future when it has resources to spare.\added{ The unfinished rule is added to a queue to be processed later.} \replaced{During rumination, the system considers the queue of unfinished rules}{It then ``ruminates" by looking at the rules currently in progress} and processes some of them so they actually fire and update the knowledge base.

An important future research goal is the implementation of the ``recipe-based planner" for Scone that would break up a goal into subgoals and plan out how to achieve each subgoal by considering different options with multiple contexts. Together with the rule-based planner, these planners would guide an intelligent agent to make decisions in order to reach a defined goal.

Simpler and more automatic ways to define rules, through structured or unstructured natural language, is an interesting direction for research. One example is a more declarative form for defining rules using structured language, such as ``the start time plus the duration of a meeting equals the end time of that meeting." The goal would be for this declarative form to create three rules, one for computing each of the start time, duration, and end time of a meeting given the other two values. There is ongoing research on automatically adding knowledge to the Scone knowledge base from external sources of unstructured natural language sentences, and this research could eventually be extended to adding production rules automatically as well.

Finally, a way to learn production rules automatically may be a desirable long-term goal, though possibly out of scope of the capabilities of Scone. Soar and ACT-R describe methods of learning production rules automatically \cite{anderson1996act, laird2019soar}, though they generally involve some external feedback or input from the environment. One method of learning from Soar is called chunking, where new production rules are learned when the system reaches an impasse and no existing production rules apply. In this case, Soar creates a new subgoal to resolve the impasse and creates a new ``chunk" rule to avoid future impasses in similar situations. In Scone, the recipe-based planner would take care of planning through such impasses, and it could implement a similar method of learning new helper production rules.

\backmatter



\bibliographystyle{plainnat}
\bibliography{bibliography} 

\appendix
\begin{appendices}
\chapter{Appendix}
We have implemented the production rule system in Common Lisp in the main Scone engine\footnote{The code for the production rule engine is currently hosted at \url{https://github.com/jchen1352/scone/tree/develop}}. There are two new user-level macros that are used to add new if-added rules and if-needed rules to the knowledge base. \texttt{new-if-added-rule} is a macro that takes bindings, x-y-z-preds, and any number of body forms as arguments. \texttt{new-if-needed-rule} takes bindings, x-y-z-preds, and an action as arguments.

Each binding is either a single variable or a list containing a variable and optional keyword arguments \texttt{:superior} and \texttt{:proper}. Each x-y-z-predicate is a three-element list where the second element is a role or relation. For if-added rules, the action is arbitrary code defined in the remainder of the body forms. For if-needed rules, the action is of the form (X Y Z) where X is a computation, Y is an individual role node, and Z is a variable defined in the bindings.

The example if-added rule described in Section \ref{if-added-example} can be defined using the macro as follows:
\vspace{3mm}
\begin{verbatim}
(new-if-added-rule (a (b :superior {airplane}))
                   ((b {travel vehicle} a))
  (new-is-a a {flying event}))
\end{verbatim}
\vspace{3mm}
Running this macro adds this rule to the knowledge base and adds triggers for this rule to \{airplane\} and \{travel vehicle\}. Assuming the Scone elements are defined correctly, we can test that the rule fires correctly with the following snippet of code:
\vspace{3mm}
\begin{verbatim}
(new-indv {my trip} {traveling event})
(new-indv {my vehicle} {airplane})
(x-is-a-y-of-z {my vehicle} {travel vehicle} {my trip})
(assert (simple-is-x-a-y? {my trip} {flying event}))
\end{verbatim}
\vspace{3mm}
This code creates a new individual trip, a new individual vehicle that is an airplane, and sets \{my vehicle\} as the \{travel vehicle\} of the trip. The assertion succeeds, indicating that the rule succesfully fired and added the link ``my trip is a flying event" to the knowledge base.

The example if-needed rule described in Section \ref{if-needed-example} can be defined using the macro as follows:
\vspace{3mm}
\begin{verbatim}
(new-if-needed-rule ((a :proper t) (b :proper t) c)
                    ((a {start time} c)
                     (b {end time} c))
                    ((scone-subtract b a) {duration} c))
\end{verbatim}
\vspace{3mm}
For simplicity, assume \{start time\} and \{end time\} are Scone numbers representing seconds since the epoch, and \texttt{scone-subtract} is an appropriately defined Lisp function that can subtract two Scone time elements. We can test that the rule fires correctly with the following snippet of code:
\vspace{3mm}
\begin{verbatim}
(new-indv {my meeting} {meeting})
(x-is-the-y-of-z {1609459200} {start time} {my meeting})
(x-is-the-y-of-z {1609462800} {end time} {my meeting})
(assert (is-x-eq-y? {3600} (the-x-of-y {duration} {my meeting})))
\end{verbatim}
\vspace{3mm}
This code creates a new individual meeting and sets its start time and end time. \texttt{the-x-of-y} is responsible for checking potential if-needed rules, and the assertion succeeds indicating that the if-needed rule successfully fired and added the link ``3600 (seconds) is the duration of my meeting" to the knowledge base.
\end{appendices}

\end{document}